# Contour Primitive of Interest Extraction Network Based on One-Shot Learning for Object-Agnostic Vision Measurement

Fangbo Qin, Jie Qin, Siyu Huang, De Xu*, *Senior Member, IEEE*

*Abstract*—Image contour based vision measurement is widely applied in robot manipulation and industrial automation. It is appealing to realize object-agnostic vision system, which can be conveniently reused for various types of objects. We propose the contour primitive of interest extraction network (CPieNet) based on the one-shot learning framework. First, CPieNet is featured by that its contour primitive of interest (CPI) output, a designated regular contour part lying on a specified object, provides the essential geometric information for vision measurement. Second, CPieNet has the one-shot learning ability, utilizing a support sample to assist the perception of the novel object. To realize lower-cost training, we generate support-query sample pairs from unpaired online public images, which cover a wide range of object categories. To obtain single-pixel wide contour for precise measurement, the Gabor-filters based non-maximum suppression is designed to thin the raw contour. For the novel CPI extraction task, we built the Object Contour Primitives dataset using online public images, and the Robotic Object Contour Measurement dataset using a camera mounted on a robot. The effectiveness of the proposed methods is validated by a series of experiments.

## I. Introduction

Vision measurement means using camera to precisely sense the spatial pose and structure of a viewed object, which is widely applied in robotic and industrial tasks [1-3]. Compared to other computer vision tasks, vision measurement focuses on geometric representation and spatial pose, instead of textured appearance, dense reconstruction, and category identification. Besides, for many robotic and industrial cases requiring high precision, the coarse visual perception is not sufficient. Therefore, geometric feature extraction is essential for vision measurement. Combining the geometric features and imaging model, 3D spatial information can be obtained [4,5].

Object contours are widely used in vision measurement. First, contour feature is more robust to partial occlusion and missing than point feature. Second, contour provides the sparse and informative geometric representation of object. In [6], circle contours were used to measure the 3D position of drogue of aerial vehicle. In [7], the pose measurement of space non-cooperative target was based on both circle and line contours [7]. For the pose alignment of high precision devices, a set of line segments on the objects' end-faces was used to reflect the spatial pose error [8]. In these works, the feature extraction methods were only suitable for the specified object types, and not scalable for novel objects.

To improve the intelligence and scalability of image contour based vision measurement, it is appealing to address the following two problems: *object-agnostic vision measurement* and *contour of interest extraction*. First, inspired by the recent works on class-agnostic vision [9-11], we attempt to explore the object-agnostic geometric feature extraction, so that a vision measurement system can be flexibly applied to various objects over different scenarios. Second, instead of extracting general contours globally, a measurement task mainly concerns a set of contours of interest that are highly related to the task purpose and have geometric meaning. Towards object-agnostic vision measurement with better reusability and scalability, this paper aims to realize the end-to-end object-agnostic contour of interest extraction. Our contribution is as follows:

1) The *contour primitive of interest extraction network* (CPieNet) is proposed based on one-shot learning, which extracts a set of pixels representing a specified *contour primitive of interest* (CPI) on an object from its raw image. One-shot learning enables the model to work on novel object by involving a support sample with annotation.

2) To obtain the one-pixel wide CPI, a *Gabor-filters based non-maximum suppression* (GF-NMS) method is proposed to thin the raw CPI output by CPieNet.

3) Because it is costly and tedious to capture and annotate numerous support-query image pairs of objects, we design an automatic sample pair generation method, to convert an annotated image to a sample pair by random transformation.

4) To the best of our knowledge, this work is the first to explore the one-shot learning of CPI extraction. For this novel task, we built the *Object Contour Primitives* (OCP) dataset using online public images, and the Robotic Object Contour Measurement (ROCM) dataset[1] using images of 15 objects collected by an eye-in-hand robot.

This work was supported by National Key Research and Development Program of China (2018AAA0103005), Science and Technology Program of Beijing Municipal Science and Technology Commission (Z191100008019004), and National Natural Science Foundation of China (61873266).

F. Qin, J. Qin, and D. Xu is with Research Center of Precision Sensing and Control, Institute of Automation, Chinese Academy of Sciences, Beijing 100190, China. {qinfangbo2013@ia.ac.cn; de.xu@ia.ac.cn}. S. Huang is with Department of Automation, Tsinghua University, Beijing, 100084, China.

---

[1] https://github.com/SURA23/Robotic-Object-Contour-Measurement-Dataset

## II. RELATED WORKS

### A. Object-agnostic Contour Based Vision Measurement

Image contour based visual measurement is preferred due to its guaranteed accuracy, robustness and sparsity. Recently, the systems that can be reused among different object types were developed. He *et al.* proposed a sparse template based 6D pose estimation method for industrial metal parts, which relies on line segment detection and cannot work on circular-shape objects [12]. In [13], the silhouette contour was extracted and used to match the nearest template, for pose estimation of textureless object, whose real-time performance was limited. In [14], the contour primitives of interest extraction (CPIE) method was proposed, which used a CPI template to match the object, then executed pixel-level analysis near the matched CPIs for precise geometric calculation. CPIE is only effective with grayscale image and highly structured scene. In comparison, CPieNet is fast and end-to-end, inferring CPIs directly from raw image. Besides, deep learning technology brings the promising generalization ability over various objects and conditions.

### B. Deep Learning Based Contour Detection

Deep learning based contour detection models outperform traditional methods, due to its powerful hierarchical feature learning ability [15]. Edge detection and boundary prediction are similar tasks [16,17]. Semantic edge detection not only extract the edge pixels but also tells which category of object each edge belongs to [18]. Line segment detection parses the line-like contours [19]. These methods presented the promising performance of deep learning on contour-related perception, but provided general low-level features which lack the task-awareness. CPieNet focuses on CPIs that have geometric meaning and are of interest to a measurement task.

### C. One/few-shot Learning for Image Perception

One/few-shot learning aims to overcome the data scarcity problem in deep learning. Especially in robotic and industrial applications, it is impracticable to build a training dataset every time a novel-type object is given. According to the recent methods, given a query image of a novel object, its perception can be helped by one or a few annotated support images of the same object, providing a *prototype* describing the object based on masked average pooling (MAP). PANet densely compared the query image's feature map with the prototype using cosine distance as metric, and the prototype alignment regularization (PAR) was used in training [20]. In [21], feature weighting was applied before dense comparison to encourage the higher feature response of foreground. CANet alternately used concatenation instead of cosine distance for dense comparison, and the iterative optimization module was designed to refine the result [10]. SG-One realized the one-shot similarity guidance, using the cosine similarity between prototype and guidance features to reweight the features in segmentation branch [22]. A-MCG used the foreground to provide guidance [23], which is not suitable for CPI extraction because CPI foreground cannot provide sufficient contextual features. Although CPI extraction is a variant of one-shot semantic segmentation, the difference between the regular-shaped narrow CPIs and the arbitrary-shaped blocky foregrounds causes that the existed methods are not ideally appropriate to CPI extraction.

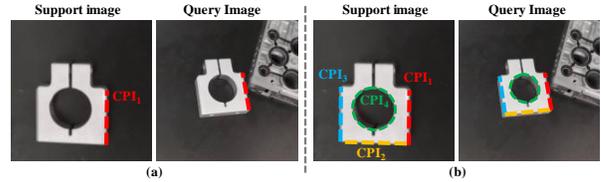

Fig. 1. Illustration of CPI extraction task. (a) Single CPI extraction. (b) Batch of CPIs extraction. CPI is manually annotated in the support image, and is extracted from the query image.

## III. PROBLEM DEFINITION

*Contour primitive* (CP) is a regular contour segment, ignoring the irregular and fragmentized contour parts. As illustrated in Fig. 1, the metal part's contour is mainly composed by several line segments and a circle. Further, CPI means a designated CP on the target object. Viewing a wide range of object categories, a majority of industrial and daily objects have the two typical CPs: *line segment* (LS) and *circular arc* (CA). In addition, LS and CA are easy for shape fitting and suitable for geometric calculation.

An image of the object to measure is regarded as the *query image* $\mathcal{I}_Q$. The task is to extract one of the CPIs on this object based on the one-shot learning CNN model $\mathcal{F}$. Assuming the object type is novel and unseen during model training, a *support image* $\mathcal{I}_S$ of the same object and its CPI annotation $\mathcal{C}_S$ are used as the guidance. $\mathcal{C}_S \in \{0,1\}$ is a binary map, whose foreground pixels mark the CPI. Thus, the CNN model is expected to extract the corresponding CPI $\mathcal{C}_Q$ from $\mathcal{I}_Q$,

$$\mathcal{C}_Q = \mathcal{F}(\mathcal{I}_Q; \mathcal{I}_S, \mathcal{C}_S) \qquad (1)$$

The task difficulty is influenced by the difference between $\mathcal{I}_S$ and $\mathcal{I}_Q$. We assume that no repeated objects occur, and the variation of imaging condition is limited, including translation, limited rotation, illumination change, color change, background change, and other stuff's occurrence. The large view-angle changes and cluttered scene are *not* involved. Fortunately, in many robotic and industrial applications, the viewed scenes are controlled. The coarse visual perception techniques can be leveraged to control the view point, region of interest, *etc.* Therefore, with the controlled difference between $\mathcal{I}_S$ and $\mathcal{I}_Q$, it is feasible to realize precise CPI extraction.

Because vision measurement usually requires multiple CPIs for geometric calculation, the single CPI extraction mode described by (1) can be easily extended to the batch of CPIs extraction mode, based on GPU's parallel computation,

$$\mathcal{C}_Q^k = \mathcal{F}(\mathcal{I}_Q; \mathcal{I}_S, \mathcal{C}_S^k) \qquad (2)$$

where $k=1,2,\ldots,N_{CPI}$. Thus, given $N_{CPI}$ query images, the CPI maps are inferred in parallel, with the shared $\mathcal{I}_S$, $\mathcal{I}_Q$ and $\mathcal{F}$.

## IV. METHODS

### A. Model Architecture

CPieNet utilizes a *support branch* to guide the *query*

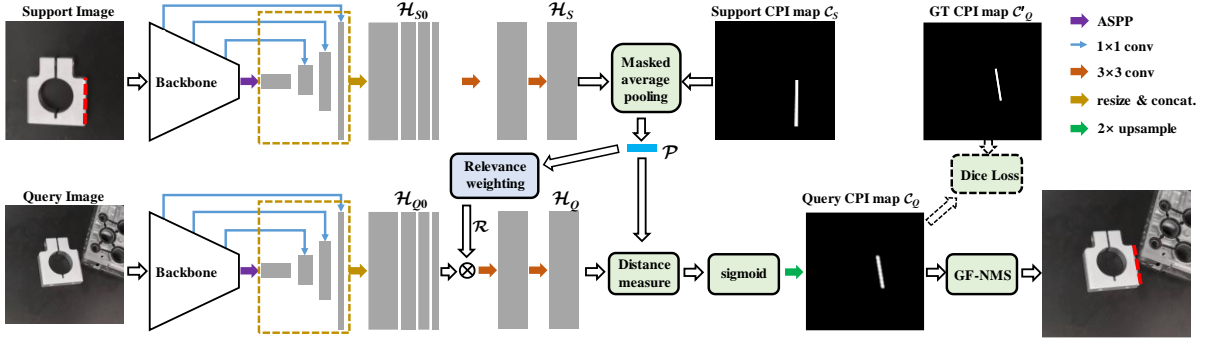

Fig. 2. CPieNet architecture. The CPI is labeled by red dashed line.

*branch*. The support branch gains the prototype vector $\mathcal{P}$ from the support image $\mathcal{I}_S$. As shown in Fig. 2, $\mathcal{I}_S$ is fed to the backbone implemented by ResNet-50 [24], and the output size is 1/16 of the input size. An atrous spatial pyramid pooling (ASPP) module [25] is used to enlarge the spatial receptive field, whose depth is 128 and atrous rates are {2,4}. The original concatenation based fusion in ASPP is replaced by the sum based fusion. The three deepest feature maps, whose sizes equal 1/2, 1/4, and 1/8 of the input size, are drawn out from backbone, then adapted to 16, 32, and 64 channels using three 1×1 convolutional layers with 1×1 stride, respectively. These adapted feature maps and the ASPP output are all resized to 1/2 of the inputs size using bilinear interpolation, and concatenated as $\mathcal{H}_{S0}$. Batch normalization is used after each of the above 1×1 convolutional layers and the ASPP's last convolutional layer, to normalize the features at different scales before fusion. The two 3×3 convolutional layers with 128 filters and 1×1 stride are used to fuse the multi-scale features in $\mathcal{H}_{S0}$, and the resulting support feature map is $\mathcal{H}_S$. Note that ReLU activation is not used in the second convolutional layer. Sharing the same backbone, convolutional layers, and their weights with the support branch, the query branch gains the multi-scale feature map $\mathcal{H}_{Q0}$ and the query feature map $\mathcal{H}_Q$ from $\mathcal{I}_Q$.

With the support feature map $\mathcal{H}_S$ and the annotated binary map $\mathcal{C}_S$, the 128-channel prototype vector $\mathcal{P}$ representing the CPI is obtained by masked average pooling,

$$\mathcal{P}_k = \frac{\sum_{i,j} \mathcal{H}_{S(i,j,k)} \times \mathcal{C}_{S(i,j)}}{\sum_{i,j} \mathcal{C}_{S(i,j)}} \quad (3)$$

where $(i, j)$ and $k$ indices position and channel, respectively.

The main guidance from support branch to query branch is based on the distance measure between $\mathcal{P}$ and the pixels on $\mathcal{H}_Q$. The cosine distance is measured by,

$$\mathcal{D}_{i,j} = \alpha \left(1 - \frac{\mathcal{H}_{Q(i,j)} \cdot \mathcal{P}}{\|\mathcal{H}_{Q(i,j)}\|_2 \times \|\mathcal{P}\|_2}\right) \quad (4)$$

where $\alpha$ is a scaling factor. Thus, $\mathcal{D}_{i,j}$ ranges from zero to $2\alpha$. Similar to [20], we set $\alpha$=20 empirically. Euclidean distance can also be used for distance measure.

After distance measure, the distance map $\mathcal{D}$ is fed to an output activation layer to obtain the query CPI map $\mathcal{C}_Q \in [0,1]$, realized using the sigmoid function,

$$\mathcal{C}_{Q(i,j)} = \frac{1}{1+e^{-(\beta-\mathcal{D}_{i,j})}} \quad (5)$$

where $\beta$ is a bias, which is set to $\beta$=5, so that the activation value $\mathcal{C}_{Q(i,j)}$ approaches 1 when $\mathcal{D}_{i,j}$ approaches zero. When $\mathcal{D}_{i,j}$ is larger than 10, $\mathcal{C}_{Q(i,j)}$ is approximately zero. Intuitively, the distance measure and sigmoid activation realize that the pixel has high output response if and only if its feature vector is close enough to the prototype vector. Finally, $\mathcal{C}_Q$ is resized back to the input size by 2× upsampling. The raw map $\mathcal{C}_Q$ is further processed by the GF-NMS module to thin the contour.

*B. Relevance Weighting*

The prototype $\mathcal{P}$ is further leveraged to reweight the multi-scale query feature map $\mathcal{H}_{Q0}$, so that the irrelevant pixels' features are pre-suppressed before the multi-scale fusion and distance measure. The weights are calculated based on the relevance $\mathcal{R}$ between $\mathcal{H}_{Q0}$ and $\mathcal{P}$,

$$\mathcal{R}_{i,j} = 1 + \frac{w_1^T \mathcal{H}_{Q0(i,j)} \cdot w_2^T \mathcal{P}}{\|w_1^T \mathcal{H}_{Q0(i,j)}\|_2 \times \|w_2^T \mathcal{P}\|_2} \quad (6)$$

where $w_1$ and $w_2$ are two learnable matrices, used to compressed $\mathcal{H}_{Q0}$ and $\mathcal{P}$ to 64 channels. $\mathcal{R}_{i,j}$ ranges from 0 to 2 as the relevance increases. Afterwards, the relevance map $\mathcal{R}$ is used to reweight $\mathcal{H}_{Q0}$ by element-wise product $\mathcal{R} \circ \mathcal{H}_{Q0}$. Thus, when a pixel $\mathcal{H}_{Q0(i,j)}$ is irrelevant to the object type, $\mathcal{R}_{i,j}$ approaches zero and this pixel's feature is suppressed.

*C. Training Loss*

At each training step, a support-query image pair $\{\mathcal{I}_S, \mathcal{I}_Q\}$ and the support CPI map $\mathcal{C}_S$ are fed into CPieNet, to predict the query CPI map $\mathcal{C}_Q$. Because CPI is narrow, the standard cross entropy (CE) loss cannot handle the pixel number imbalance between foreground and background. The weighted CE loss has a hyper-parameter weight to tune. The Dice loss is used to supervise the learning of CPI extraction, which realizes sharp contour prediction and has no extra hyper-parameter [26],

$$\mathcal{L} = 1 - \frac{\sum_{i,j} 2\mathcal{C}_{Q(i,j)} \mathcal{C}'_{Q(i,j)}}{\sum_{i,j} \mathcal{C}^2_{Q(i,j)} + \sum_{i,j} \mathcal{C}'^2_{Q(i,j)} + \tau} \quad (7)$$

where $\mathcal{C}'_Q$ is the ground-truth of query CPI map. The small positive constant $\tau$ is used to stabilize the computation.

## D. Support-query Sample Pair Generation

Since support-query sample pairs are required to train CPieNet, instead of collecting an image pair for every object and labelling them coordinately, we collect the online public images, which covers a wider range of object types with a much lower cost. Each image is annotated individually, then used to generate a sample pair automatically.

An $H{\times}W$ image and one of its CPI annotations are called a raw sample $\{\mathcal{I}_R,\mathcal{C}_R\}$. To generate a support-query sample pair from a raw sample, we customized random data augmentation to mimic imaging condition variation, as implemented by the following steps. For the convenience, the default parameters are directly presented here, which can be adjusted in practice.

1) *Mix-up*: $\mathcal{I}_R$ is mixed with another image $\mathcal{I}_1$ by weighted sum $\mathcal{I}_{mix}=(1-\gamma_{mix})\mathcal{I}_R+\gamma_{mix}\mathcal{I}_1$, where $\gamma_{mix}\in[0,0.3]$ is random. *Thus random shade is overlapped on the object.*

2) *Cutout&Patch*: Cutout a patch $\mathcal{I}_{P2}$ randomly from another image $\mathcal{I}_2$. Resize $\mathcal{I}_{P2}$ to be smaller and put in $\mathcal{I}_{mix}$ randomly without covering the CPI, *to mimic a nearby stuff.*

3) *Pad*: Select another image $\mathcal{I}_3$, center-crop a patch from it, and resize the patch to $1.4H{\times}1.4W$, to get $\mathcal{I}_{pad}$. Then $\mathcal{I}_{mix}$ is overlaid at the center of $\mathcal{I}_{pad}$, *to mimic the surrounding change*. Meanwhile $\mathcal{C}_R$ is padded to $1.4H{\times}1.4W$ with zeros.

4) *Data augmentation*: The ordinary data augmentation is used to *mimic the translation, rotation, scaling, illumination change and color change*. Thus, $\mathcal{I}_{pad}$ and $\mathcal{C}_R$ are transformed to $\mathcal{I}_{aug}$ and $\mathcal{C}_{aug}$, respectively.

5) *Crop*: $\mathcal{I}_{aug}$ and $\mathcal{C}_{aug}$ are cropped randomly to the original size of $H{\times}W$. The CPI in $\mathcal{C}_{aug}$ should not be cropped off.

6) Repeat Steps 1-5 twice, to produce $\{\mathcal{C}_S,\mathcal{I}_S\}$ and $\{\mathcal{C}_Q,\mathcal{I}_Q\}$.

## E. Gabor-Filters based Non-Maximum Suppression

The ideal contour for vision measurement should be single-pixel wide. However, the raw CPI given by CPieNet is sharp but not guaranteed single-pixel wide, because the convolution operation leads to diffused edge. In [16] and [27], non-maximum suppression (NMS) is applied along the edge's normal direction, which is estimated by the local gradient, to sharpen the raw contour. Comparing to the ordinary edge and contour that have many irregular or curly parts, the CPI in our task has regular shape, either LS or CA. Therefore, we proposed Gabor-filters based NMS (GF-NMS) to improve the contour thinning performance for CPI. Gabor filter is featured by its sensitivity to direction and spatial frequency [28]. The Gabor kernel $g$ is determined by the five parameters: standard deviation $\sigma_g$, normal direction $\theta_g$, wavelength $\lambda_g$, aspect ratio $\gamma_g$, and phase offset $\psi_g$. By selecting proper parameters and the four directions $\theta_g=\{0°,45°,90°,135°\}$, the four truncated Gabor kernels $\{g_1, g_2, g_3, g_4\}$ with the $9{\times}9$ size and ridge-shape are constructed, as visualized in Fig. 3(a).

The proposed GF-NMS is described in Algorithm 1. As illustrated in Fig. 3, after the Gabor filtering running on GPU, the four-direction response maps are obtained. For each pixel, the direction with the strongest response is regarded as the approximate normal direction of the local contour. With the direction map, NMS is conducted within the 8-pixel neighborhood along the approximate normal direction.

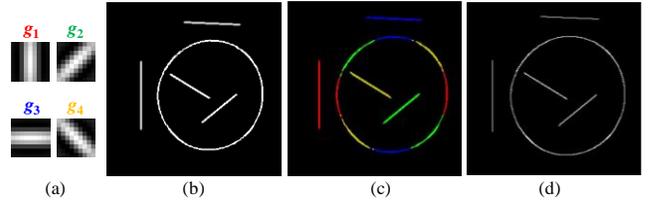

(a) (b) (c) (d)

Fig. 3. Gabor-filters based NMS. (a) Gabor kernels. (b) Raw contour. (c) Normal direction map. The directions corresponding to the 4 Gabor kernels are shown by the 4 colors, respectively. (d) Thinned contour.

---

**Algorithm 1: Gabor-Filters based NMS**

**Input:** Contour map $\mathcal{C}$, Gabor kernels $\{g_k\}$, threshold $g_0$.
**Output**: Thinned contour map $\mathcal{C}_T$.

1. Smooth: $\mathcal{S}\leftarrow g_b*\mathcal{C}$; # $g_b$ is a Gaussian kernel;
2. Initialize $\mathcal{D}$ and $\mathcal{C}_T$ with zeros;
3. **for each** $g_k$ ($k$=1,2,3,4), # *4-direction Gabor filtering*
    $\mathcal{G}_k\leftarrow g_k*\mathcal{S}$;
4. **for each** pixel $\mathcal{D}_{i,j}$ of $\mathcal{D}$, # *Obtain direction map*
    $\mathcal{D}_{i,j}\leftarrow\mathrm{argmax}(g_0, \mathcal{G}_{1i,j}, \mathcal{G}_{2i,j}, \mathcal{G}_{3i,j}, \mathcal{G}_{4i,j})\in\{0,1,2,3,4\}$;
    **if** $\mathcal{C}_{i,j}$=0, **then** $\mathcal{D}_{i,j}\leftarrow0$;
5. **for each** pixel $\mathcal{C}_{T(i,j)}$ of $\mathcal{C}_T$, # *NMS*
    **if** $\mathcal{D}_{i,j}$=1 **and** $\mathcal{S}_{i,j}\geq\mathrm{max}(\mathcal{S}_{i,j-1},\mathcal{S}_{i,j+1})$, **then** $\mathcal{C}_{T(i,j)}\leftarrow\mathcal{C}_{i,j}$;
    **if** $\mathcal{D}_{i,j}$=2 **and** $\mathcal{S}_{i,j}\geq\mathrm{max}(\mathcal{S}_{i-1,j-1},\mathcal{S}_{i+1,j+1})$, **then** $\mathcal{C}_{T(i,j)}\leftarrow\mathcal{C}_{i,j}$;
    **if** $\mathcal{D}_{i,j}$=3 **and** $\mathcal{S}_{i,j}\geq\mathrm{max}(\mathcal{S}_{i-1,j},\mathcal{S}_{i+1,j})$, **then** $\mathcal{C}_{T(i,j)}\leftarrow\mathcal{C}_{i,j}$;
    **if** $\mathcal{D}_{i,j}$=4 **and** $\mathcal{S}_{i,j}\geq\mathrm{max}(\mathcal{S}_{i+1,j-1},\mathcal{S}_{i-1,j+1})$, **then** $\mathcal{C}_{T(i,j)}\leftarrow\mathcal{C}_{i,j}$;
6. **return** $\mathcal{C}_T$.

---

## V. EXPERIMENTS

### A. OCP Dataset and Sample Pair Generation

We collected 2307 online public images containing various object types, including mechanical parts, digital products, industrial devices, household items, containers, *etc*. Each image was resized to 320×320, and had at least one CPI. 1807 images that had 4844 LS samples and 622 CA samples were used for training, and the remaining 500 images that had 1297 LS samples and 186 CA samples were used for testing.

The proposed method in Section IV.*D* was executed to transform the raw samples to sample pairs. Note that the training sample pairs were generated online at each training step, using the randomness to cover more condition variances. The test sample pairs were produced and fixed, for the fair evaluation of different methods. The ground-truth CPIs in training and test sets were 3-pixel and 1-pixel wide, respectively. The thicker CPIs were used for training because manual annotation might have slight error. The ordinary data augmentation is implemented with the *imgaug* library (https://github.com/aleju/imgaug). Affine transformation was applied, including scaling, translation, in-plane rotation within [-15°,15°], shearing within [-15°,15°], and aspect ratio changes. Then, coarse dropout and the slight changes on color were conducted. Four examples of sample pair generation are shown in Fig. 4. The proposed generation method provided additional variations. As shown in Fig. 4(b), the generated query image had both the overlapping shade and another bottle stuff near the object. As shown in Fig. 4(a), the generated query image had different background near the boundary.

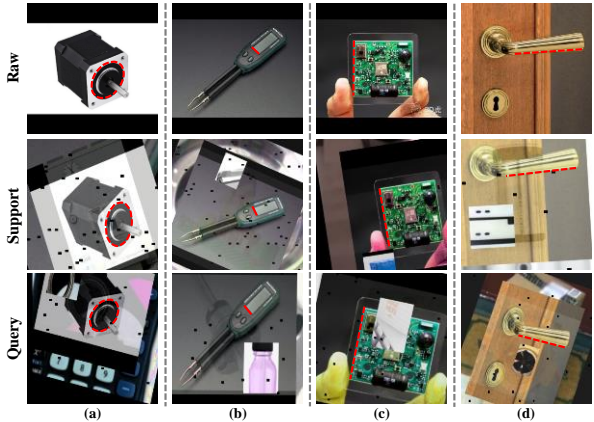

Fig. 4. Examples of support-query sample pair generation. The three rows show the annoated raw sample, generated support sample and generated query sample, respectively. CPI annotation is marked by red.

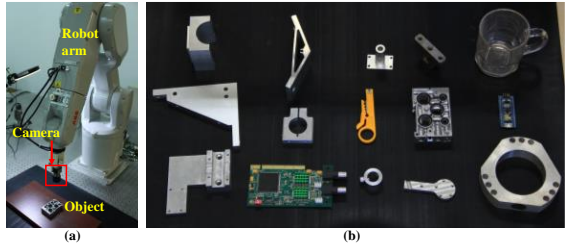

Fig. 5. Platform and objects to construct ROCM dataset.

### B. ROCM Dataset

To evaluate in the real environment, we collected the images of fifteen 3D objects with an ABB IRB-1200 robot and a Basler acA2440-35uc industrial camera with an 8mm lens, as shown in Fig. 5. For each object, the robot moved to capture the images of it from different viewpoints. Sometimes the illumination was changed and other stuffs were put near the object. The images were resized to 320×320. Thus, 15 image series including 523 images in total were obtained and annotated, providing 2188 LS samples and 334 CA samples.

Two evaluation modes were used. The first mode used the *1st frame* of an image series as the support image, and the rest frames as the query images. The second mode alternatively used a *template* image as support image, which was captured by a consumer-grade camera when putting the object on a black pad. Apparently, the template based evaluation is more challenging because the different imaging device and scene.

### C. Training Details and Evaluation Metrics

Only the OCP training set was used to train the models. The input size was 320×320. The model training is based on the Adam optimizer, with the initial learning rate of 0.0001, the batch size of 4, and the training epochs of 40. The learning rate was decayed by 0.5 every 10 epochs. The ResNet-50 backbone was pre-trained on ImageNet. Before the online sample pair generation at each training step, the image was randomly flipped vertically and horizontally. In the GF-NMS algorithm, the 5×5 Gaussian kernel had the standard deviation of 1.0 and the threshold $g_0$ was 2.0. The four Gabor kernels had the parameters $\sigma_g$=2.0, $\lambda_g$=9.0, $\gamma_g$=0.3, and $\psi_g$=0. The hardware configuration included a 3.70GHz Intel i7-8700K CPU and two NVIDIA RTX2080ti GPUs.

The OCP test set and the entire ROCM dataset were used to evaluate the models. Following the edge detection work [18], the maximum F-Measure at optimal dataset scale (MF-ODS) was adopted as the metric of CPI extraction performance, regarding CPI map as edge map. The misalignment tolerance threshold was set to 0.015×$L$, and $L$ is the diagonal length of the map. When evaluating on ROCM dataset under the template mode, the support and query images are preprocessed by illumination normalization, $V=V-V^*+127$, where $V^*$ is the Gaussian blurring based illumination map of the third channel $V$ of HSV image, to reduce the contrast differences between the support and query images.

### D. Ablation Experiments and Visualization

A series of ablation experiments were conducted and the results are reported in Table I. The experiment No. 1 was regarded as the baseline, which used the CPieNet with cosine distance but without relevance weighting (RW). 1) In No. 2 experiment, Euclidean distance led to the worse performance. 2) In No. 3 experiment, using only the ordinary data augmentation in Step 4 of Section IV.*D*, the performance degraded significantly. 3) In No. 4 experiment, the raw contour was directly used for evaluation without using GF-NMS, resulting in the lower scores, which showed the necessity of contour thinning. 4) In No. 5 experiment, with the proposed RW, the scores on ROCM dataset were increased. Besides, CPieNet presented the real-time speed with the 320×320 input size.

In Fig. 6, two examples of CPieNet inference on OCP dataset are visualized. The relevance maps show that the RW module could automatically learn to highlight the pixels more relevant to the target CPI. With the relevant pixels' features enhanced and the irrelevant pixels' features suppressed, the following distance measure can be more concentrated on the discrimination of similar but different contour parts. As shown in the 2nd row, CPieNet without RW failed to distinguish the line segments with and without screw thread.

We visualize the batch of CPIs extraction results with CPieNet on ROCM dataset in Fig. 7. As shown in the first row, CPieNet extracted the four CPIs, which could be used to localize the silicon chip. The overall experiments revealed that CPieNet somtimes failed when background or view angle changed significantly, because the generated training sample pairs could not cover all the variations in real evironment.

### E. Comparison Experiments

The related one/few-shot segmentation methods [10,20-22] were re-implemented for comparison. All the methods shared the same loss function, training configuration, GF-NMS, MAP based prototype, and feature extractor to obtain $\mathcal{H}_S$ and $\mathcal{H}_Q$. 1) *CANet*: $\mathcal{H}_Q$ and $\mathcal{P}$ are concatenated, then fused by a 3×3 convolutional layer with 128 channels and the dilated rate of 2. The intermediate convolutional layers in the iterative optimize module all had 128 channels, and the iterative refinement was repeated by 4 times. 2) *PANet* and *PANet-sigmoid*: The original PANet used both the foreground and background prototypes, and its output is based on softmax function. PANet-sigmoid used only the foreground prototype and the sigmoid based output given by Eq. (4). 3) *SG-One*: The guidance branch had three 3×3 convolutional layers with 128 channels and the 1×1 stride. The segmentation branch was formed by 1×1 convolutional layers with 128 channels.

TABLE I
ABLATION EXPERIMENTS ON CPIENET

| No. | MF-ODS | | | Time (ms) |
|---|---|---|---|---|
| | OCP | ROCM (1st frame) | ROCM (template) | |
| 1. Cosine distance (baseline) | **0.827** | 0.850 | 0.825 | 16.5 |
| 2. Euclidean distance | 0.790 | 0.807 | 0.783 | 16.4 |
| 3. Simple sample pair generat. | 0.772 | 0.827 | 0.800 | 16.5 |
| 4. without GF-NMS | 0.694 | 0.695 | 0.678 | 15.8 |
| 5. with relevance reweighting | 0.826 | **0.856** | **0.828** | 17.4 |

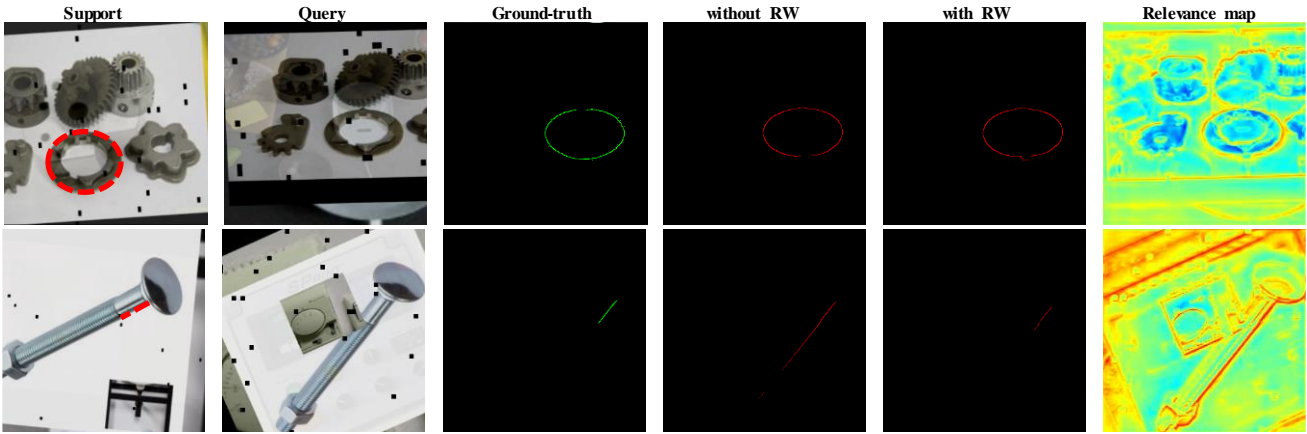

Fig. 6. Single CPI extraction on OCP dataset. The 1st column shows the support images and the CPI annotations (red dashed lines). In the relevance maps, blue and red indicate the lowest and highest values, respectively.

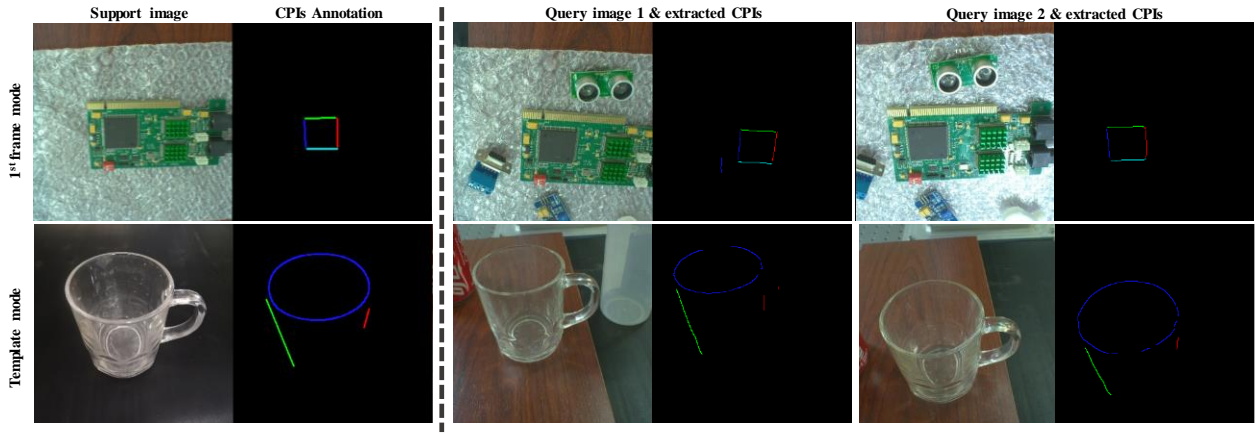

Fig. 7. Batch of CPIs extraction on ROCM dataset. Each row show an example. Different colors mark the multiple different CPIs on the same object.

TABLE II
COMPARISON EXPERIMENTS

| Method | MF-ODS | | | Time (ms) |
|---|---|---|---|---|
| | OCP | ROCM (1st frame) | ROCM (template) | |
| CANet [10] | 0.820 | 0.814 | 0.789 | 24.2 |
| PANet [20] | 0.752 | 0.777 | 0.758 | 16.6 |
| PANet-sigmoid [20] | 0.825 | 0.845 | 0.825 | 16.4 |
| Feature Weighting [21] | 0.806 | 0.823 | 0.802 | 17.4 |
| SG-One [22] | 0.776 | 0.827 | 0.801 | 18.7 |
| CPieNet+NMS [16, 27] | **0.832** | 0.848 | **0.832** | 19.7 |
| CPieNet | 0.826 | **0.856** | 0.828 | 17.4 |

The evaluation results on the two dataset are reported in Table II. PANet-sigmoid was superior than PANet because the background prototype used in PANet might have poor generalization ability when the background changes. In comparison, CPieNet demonstrated the best overall performance on the two datasets. Besides, we investigated the contour thinning performances of the gradient-based NMS [16,27] and the proposed GF-NMS. GF-NMS presented the comparable accuracy and cost less runtime because its main computation was on GPU.

## VI. CONCLUSION

Object-agnostic geometric feature extraction is an essential step to realize object-agnostic vision measurement. Towards this target we propose the CPieNet model under the one-shot learning framework. Given an image of a novel-type object, CPieNet extracts the designated CPI from it, according to the prototype obtained from an annotated example. The relevance weighting module is embedded to improve the discrimination ability by enhancing relevant pixels before dense similarity comparison. GF-NMS is proposed to thin the CPI to one-pixel wide, considering the requirement of precise measurement. The paired training samples are generated from online public images, with lower cost and wider range of object types. The two novel datasets OCP and ROCM are built for training and evaluating the proposed model. The future work will continue to improve the robustness and accuracy of CPI extraction.